\documentclass{article}

\usepackage{arxiv}

\usepackage[utf8]{inputenc} % allow utf-8 input
\usepackage[T1]{fontenc}    % use 8-bit T1 fonts
\usepackage{hyperref}       % hyperlinks
\usepackage{url}            % simple URL typesetting
\usepackage{booktabs}       % professional-quality tables
\usepackage{amsfonts}       % blackboard math symbols
\usepackage{nicefrac}       % compact symbols for 1/2, etc.
\usepackage{microtype}      % microtypography
\usepackage{graphicx}
\usepackage{xcolor}    
\usepackage{natbib}
\usepackage{doi}
\usepackage{multirow}    % 合并单元格功能（可选）
\usepackage{array}       % 自定义列宽
\usepackage{caption}
\usepackage{amsmath}
\usepackage{algorithm}
\usepackage{algpseudocode}
\usepackage{wrapfig}

\title{ReconMOST: Multi-Layer Sea Temperature Reconstruction with Observations-Guided Diffusion}

\author{
\normalsize \textbf{Yuanyi Song}\textsuperscript{1},
\normalsize ~\textbf{Pumeng Lyu}\textsuperscript{2}, 
\normalsize ~\textbf{Ben Fei}\textsuperscript{3,*}, \\
\normalsize \textbf{Fenghua Ling}\textsuperscript{2}, 
\normalsize \textbf{Wanli Ouyang}\textsuperscript{3}, 
\normalsize \textbf{Lei Bai}\textsuperscript{2}\\
\normalsize \textsuperscript{1}Shanghai Jiao Tong University, \normalsize \textsuperscript{2}Shanghai AI Laboratory, \normalsize \textsuperscript{3}The Chinese University of Hong Kong\\
\normalsize \texttt{norsheep919@sjtu.edu.cn, benfei@cuhk.edu.hk} \\
\textsuperscript{*} Corresponding Author
}
\date{}

\renewcommand{\shorttitle}{\textit{arXiv} Template}

\hypersetup{
pdftitle={A template for the arxiv style},
pdfsubject={q-bio.NC, q-bio.QM},
pdfauthor={David S.~Hippocampus, Elias D.~Striatum},
pdfkeywords={First keyword, Second keyword, More},
}

\begin{document}
\maketitle

\begin{abstract}
% \lipsum[1]
Accurate reconstruction of ocean is essential for reflecting global climate dynamics and supporting marine meteorological research. Conventional methods face challenges due to sparse data, algorithmic complexity, and high computational costs, while increasing usage of machine learning (ML) method remains limited to reconstruction problems at the sea surface and local regions, struggling with issues like cloud occlusion.
To address these limitations, this paper proposes ReconMOST, a data-driven guided diffusion model framework for multi-layer sea temperature reconstruction. Specifically, we first pre-train an unconditional diffusion model using a large collection of historical numerical simulation data, enabling the model to attain physically consistent distribution patterns of ocean temperature fields. During the generation phase, sparse yet high-accuracy in-situ observational data are utilized as guidance points for the reverse diffusion process, generating accurate reconstruction results. Importantly, in regions lacking direct observational data, the physically consistent spatial distribution patterns learned during pre-training enable implicitly guided and physically plausible reconstructions. 
Our method extends ML-based SST reconstruction to a global, multi-layer setting, handling over 92.5\% missing data while maintaining reconstruction accuracy, spatial resolution, and superior generalization capability. 
We pre-train our model on CMIP6 numerical simulation data and conduct guided reconstruction experiments on CMIP6 and EN4 analysis data. The results of mean squared error (MSE) values achieve 0.049 on guidance, 0.680 on reconstruction, and 0.633 on total, respectively, demonstrating the effectiveness and robustness of the proposed framework. Our source code is available at \url{https://github.com/norsheep/ReconMOST}.
\end{abstract}

% keywords can be removed
% \keywords{First keyword \and Second keyword \and More}

% \input{Sections/Introduction}

% \input{Sections/Related}

% \input{Sections/Method}

% \input{Sections/Experiments}

% \input{Sections/Conclusion}
\section{Introduction}
\label{sec:introduction}
The ocean, covering around 70\% of Earth's surface, can absorb and release large amounts of heat and thus stabilize Earth's climate~\citep{johnson2016improving, von2016imperative}. Tremendous efforts have been made to monitor ocean temperature, but still with irregular and sparsely distributed observations from Argo floats, ships and satellites~\citep{roemmich2009argo,abraham2013review, von2013monitoring}. Sea temperature (ST) reconstruction is thus needed for a more complete picture of ocean temperature and different depths and locations. It has a significant impact to help us understand global warming~\citep{mendelssohn2012reanalysis}, discover factors driving extreme weather~\citep{jewson2007predicting,prochaska2021deep}, and understand climate dynamics~\citep{cutolo2024cloinet}.

Conventional approaches for climate field reconstruction mostly include Optimal Interpolation (OI)~\citep{smith2003extended}, Kriging~\citep{youzhuan2008reconstruction}, and Empirical Orthogonal Functions (EOF)~\citep{alvera2005reconstruction}.
% However, these statistical methods exhibit limitations, including lower reconstruction accuracy in large and irregular areas of missing data, high computational costs, and limited generalization capability across the various ocean domains. 
However, these data science methods exhibit limitations, for example, limited spatial coverage, lower reconstruction accuracy in large and irregular areas of missing data, high computational costs, and limited generalization capability across the various ocean domains~\citep{smith2007objective,zhang2022gdcsm_argo}.

\begin{figure}[t]
    \centering
    \includegraphics[width=1.0\columnwidth]{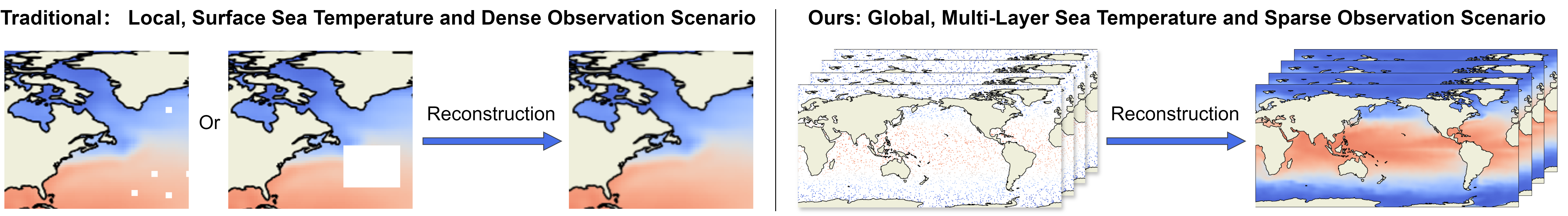}
    \vspace{-0.5cm}
    % \caption{Main Different between Traditional Method and Ours ReconMOST.}
    \caption{Difference between traditional methods and our ReconMOST.}
    % \vspace{-0.1cm}
    \label{fig:compare}
\vspace{-0.3cm}
\end{figure}

Recent development of AI has shown promising solutions for ocean reconstruction. Neural networks and generative AI models are used to reconstruct sea surface temperature (SST)~\citep{ouala2018sea} and understand ocean dynamics~\citep{zheng2024generating}.
These studies indicate that ML-based methods have the potential to outperform traditional statistical approaches in ocean reconstruction. However, these models are still used for dense observation or cloud occulasion scenarios in the local sea, shown as the left part of Fig.~\ref{fig:compare}, and often fail in scenarios with extreme data sparsity, such as global sea temperature reconstruction, where observations (e.g., from Argo floats or satellites) are significantly fewer than required grid points. This sparsity complicates learning the underlying physical patterns, particularly for multi-layer ocean temperature fields. To our knowledge, no existing work has addressed the reconstruction of global, deep ocean temperature fields under such sparse conditions, a critical task for advancing climate modeling and oceanographic research.

To address these challenges, we propose ReconMOST, a novel model based on the Guided Diffusion Model~\citep{dhariwal2021diffusion} for reconstructing near-globally distributed, multi-layered sea temperature fields from extremely sparse observational data shown as the right part of Fig.~\ref{fig:compare}, demonstrating strong generalization capability to real-world observations.
~\citet{alessio1999space} and Kawamura~\citep{kawamura1994rotated} have indicated that the spatial distribution of ocean temperature fields exhibits consistent principal patterns across varying temporal scales.
Inspired by these findings, we first trained an unconditional diffusion model using a large collection of historical ocean temperature data from numerical model simulations. 
Using the pre-trained diffusion model, we guide the reverse generation process to reconstruct target multi-layer sea temperature fields under real-world sparse observation conditions. 
While ensuring accurate reconstruction at the observational guidance points, the physically consistent distribution patterns learned by the pre-trained diffusion model implicitly enhance the reconstruction quality of unobserved regions as well. 
Experimentally, a diffusion model was trained on seven numerical model's historical simulation data, which effectively learns a physically consistent spatial prior for 3D ocean temperature fields. We then conduct reconstruction experiments on both remaining CMIP6 and EN4 analysis data, and ensure that the number of guidance points matches the buoy data density. 
Our method achieves a mean squared error (MSE) of 0.633. Additionally, we conduct ablation experiments to evaluate the effects of different pre-training modes, dataset sizes, and guidance parameters, providing robust evidence of our method's effectiveness in various aspects. Our main contributions are threefold:
\begin{itemize}
\vspace{-0.2cm}
\setlength{\itemsep}{0pt}
\setlength{\parskip}{0pt}
    \item \textbf{We introduce a data-driven guided diffusion framework for ocean reconstruction for the first time.} Our ReconMOST captures physically consistent spatial patterns through large-scale numerical simulations prior, enabling the reliable generation of ocean states aligned with fundamental physical properties.
    \item \textbf{Our framework reconstructs global, multi-layer sea temperature fields under extreme sparsity, with up to 92.5\% missing data.} Even in such scenarios, the method maintains a high reconstruction accuracy, achieving an MSE as low as 0.633, demonstrating strong robustness in sparse-data regimes.
    \item \textbf{We simultaneously achieve high reconstruction accuracy and generalization performance.} Our ReconMOST adapts to various densities of sparse observational points according to real-world observations. As the number of guided observations increases, reconstruction accuracy improves correspondingly, while retaining reliable performance under minimal supervision.
\end{itemize}

\section{Related Work}
\label{sec:related}
\textbf{Guided Diffusion.}
Guided diffusion is a conditional sampling technique for diffusion models, where external information is incorporated by computing the gradient between the generated sample and the conditioning data, and using this gradient to steer the sampling trajectory~\citep{tu2024taming,lyu2024getmesh}. % This approach has been widely applied across various domains, particularly in image-related tasks.\citep{wang2022guided} 
This approach has been widely applied across various domains. ~\citet{wang2022guided}
mitigates adversarial perturbations in image classification, while SGDiff and GDUI ~\citep{sun2023sgdiff,xie2024gdui} focus on style-controlled image synthesis and unsupervised image generation, respectively. These methods leverage essential latent features of the distribution while matching relevant conditions. Beyond image, guided diffusion has been extended to other tasks by reframing them as a structured image processing problem. For instance, DMDiff ~\citep{zhang2025dmdiff} addresses cloud removal by fusing remote sensing data, and ~\citet{klarner2024context} handle out-of-distribution molecular and protein generation. Inspired by these advances, sea temperature reconstruction from sparse observations can similarly be formulated as a conditional large-scale image inpainting task, where the goal is to recover a physically consistent, multi-layered ocean field conditioned on limited point-wise measurements.

\textbf{Ocean Reconstruction. }
Ocean field reconstruction has been constrained by the lack of accurate ground truth data, primarily due to observational errors and the inherent discrepancies between physical simulations and real-world ocean conditions. Existing methods predominantly rely on numerical outputs or reanalysis products from authoritative institutions, such as the ECMWF or the WCRP, as reference datasets~\citep{hersbach2016era5, gutowski2016wcrp}. Consequently, most data-driven ocean and atmospheric reconstruction approaches are trained on these numerical model outputs.
Recent advances in AI, particularly neural networks, offer promising solutions. \citet{ouala2018sea} first used deep learning to reconstruct sea surface temperature (SST) by introducing bilinear residual neural network representations, which mimic numerical integration schemes such as Runge-Kutta~\citep{cartwright1992dynamics}, for the forecasting and assimilation of geophysical fields from satellite-derived remote sensing data. ~\citet{zheng2024generating} redesigned a purely data-driven latent space DA framework (DeepDA) that employs a generative AI model to capture the nonlinear evolution in sea surface temperature. Despite their potential to outperform traditional statistical methods, these ML approaches do not work well in scenarios with extreme data sparsity, where limited observations hinder the learning of underlying physical patterns.
  
\textbf{Guided Diffusion in Reconstruction Issues.}
In atmospheric sciences, forecasting and field reconstruction problems typically produce spatially discrete results, which can naturally be reframed as conditional image generation or restoration tasks. Guided diffusion models, with their inherent capability for conditional sampling, have demonstrated strong adaptability in this context. SGD~\citep{tu2025satellite} integrates conditional diffusion models with attention mechanisms to reconstruct high-resolution meteorological states from low-resolution inputs. ~\citet{hua2024weather} incorporates numerical weather prediction outputs during sampling, enabling flexibility across varying forecast lead times and improving long-range forecast stability. Similarly, DGDM~\citep{yoon2023deterministic} uses deterministic forecasts to guide probabilistic diffusion models, achieving high-precision probabilistic predictions.
These advances collectively demonstrate the effectiveness of guided diffusion models in ocean reconstruction tasks, highlighting their potential for addressing the highly sparse and multi-layer global sea temperature field reconstruction.

\section{Method}
\label{sec:method}
\begin{figure}[t]
    \centering
    \includegraphics[width=1.0\columnwidth]{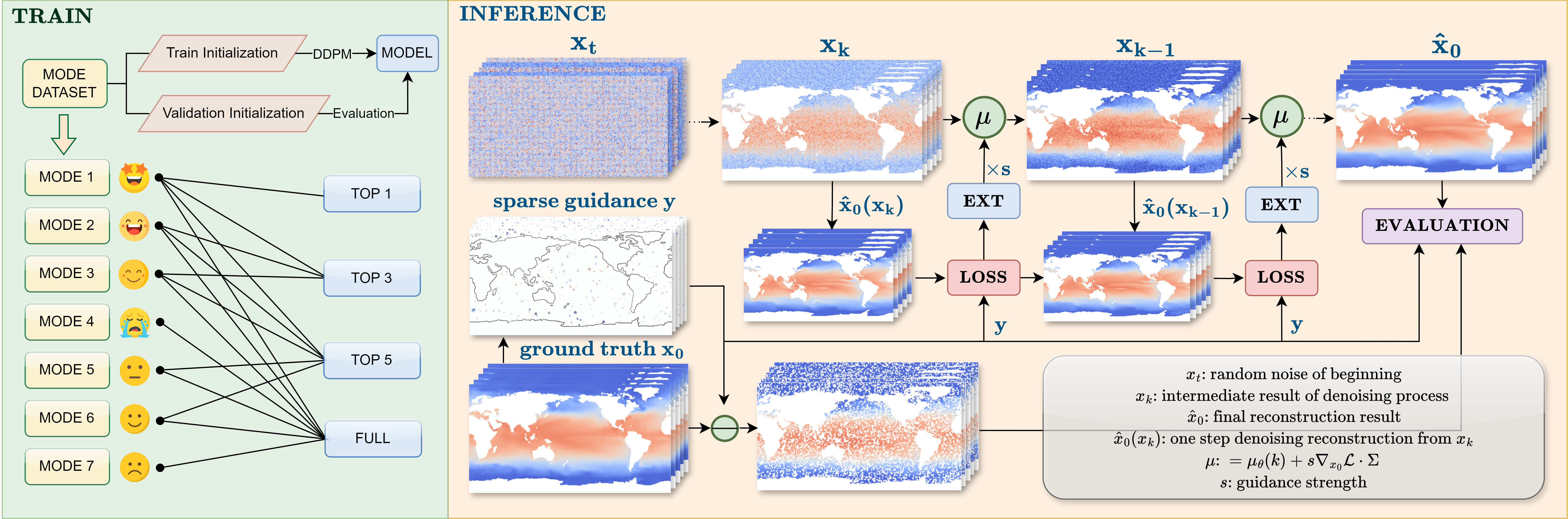}
    \caption{The framework of ReconMOST: In the training phase, we pre-train an unconditional diffusion model using a large collection of historical numerical simulation data chosen by effectiveness experiment in Sec.~\ref{sec:experiments_effectiveness}, enabling the model to learn physically consistent distribution patterns of ocean temperature fields.
    In the inference phase, we split the test dataset into sparse guidance part (7.5\%), which is utilized to guide the reverse process to generate accurate reconstruction results, and evaluation part, which is utilized for the evaluation the accuracy of reconstruction results in Sec.~\ref{sec:experiments_generalization}.}
    %\vspace{-0.8cm}
    \label{fig:pipeline}
\vspace{-0.5cm}
\end{figure}

Our ReconMOST framework follows a train-inference process as Fig.~\ref{fig:pipeline}. 
First, we train a diffusion model on a large dataset of historical ocean temperature profiles derived from numerical simulations. This process allows the model to learn the underlying spatial distribution patterns of sea temperature fields across both horizontal and vertical dimensions. 
Second, during the reconstruction phase, sparse observational data of both surface and subsurface measurements are incorporated as guidance. 
These observations dynamically guide the reverse sampling trajectory of the diffusion process, enabling the model to reconstruct multi-layer sea temperature fields that not only match the provided observations but also preserve the learned global statistical distribution with physical consistency of the ocean temperature field.
% \vspace{3pt}

\subsection{Pretrain Physically Consistent Diffusion Model}

The physically consistent diffusion model is obtained by using the classical denoising diffusion probability model (DDPM)~\citep{ho2020denoising}. 
Specifically, we pre-train the model using a large corpus of historical ocean temperature profile data generated by numerical simulations. It is well-known that ocean temperature fields exhibit fundamental and stable physical distribution patterns on a global scale. For example, sea surface temperatures are typically higher near the equator and decrease progressively toward the polar regions~\citep{NOAA2022}. 
Additionally, due to the effects of oceanic circulation, different ocean basins exhibit distinct temperature characteristics~\citep{NOAA2024}. For instance, the surface temperatures of the Pacific Ocean 
\begin{wrapfigure}{r}{0.45\textwidth}
\vspace{-10pt}
\begin{minipage}{0.45\textwidth}
\begin{algorithm}[H]
\renewcommand{\baselinestretch}{1.2}\selectfont % 设置行距1.2倍
\caption{\textbf{Observation Guided Reconstruction}}
\label{alg:guided-sample}
\begin{algorithmic}[1]  % [1] 表示行号起始
    \State \textbf{Input:} Sparse observation data $y$, gradient scale $s$, distance measure $\mathcal{L}$, Gaussian Kernal $\mu,\sigma$.
    \State \textbf{Output:} Output image $x_0$ conditioned on $y$
    \State Sample $x_T$ from $\mathcal{N}(0, \mathbf{I})$
    \For{$t$ from $T$ to $1$}
        \State $\hat x_0 = \frac{x_t - \sqrt{1-\alpha_t} \epsilon_\theta(x_t, t)}{\sqrt{\alpha_t}}$
        \State $g = -s\cdot\nabla_{x_0}L(\hat x_0,y)$
        \State $g=\text{SoftExtension}(g,\mu,\sigma)$
        \State Sample $x_{t-1} \sim \mathcal{N}\left(\mu + \Sigma g, \Sigma \right)$
    \EndFor
    \State \Return $x_0$
\end{algorithmic}
\end{algorithm}
\end{minipage}
\vspace{-10pt}
\end{wrapfigure}
are generally higher than those of the Atlantic, and the deep-water masses in various basins also follow corresponding distribution patterns.
These statistical characteristics, derived from extensive historical observations and numerical simulations, imply the underlying physical laws governing the distribution of ocean temperatures. Such basic distribution can be modeled by DDPM, effectively capturing the spatial distribution patterns and physical consistency of ocean temperature fields. This ensures that the subsequent reconstruction process produces physically plausible results that conform to both observational constraints and large-scale statistical regularities.

% The core principle of DDPM is learning the data distribution and its transformation process through a sequential noise-adding and denoising procedure. Accordingly, we adopt DDPM as the foundation model for our model pretraining, enabling it to effectively capture the spatial distribution patterns and physical consistency of ocean temperature fields. This ensures that the subsequent reconstruction process produces physically plausible results that conform to both observational constraints and large-scale statistical regularities.

% 这一段是想先简单概括一下DDPM的训练过程
% DDPM progressively transforms $x_0 \sim p_{\text{data}}$, into samples from a simple Gaussian noise distribution $x_T\sim\mathcal{N}(0,1)$. By iteratively predicting the noise added to original sample $x_0$ to get $x_t,t\in[0,T]$ at each discrete time, and gradually reversing the process to recover $x_0$. The result at the end of this reverse denoising process corresponds to a sample from the learned data distribution $p_{\text{data}}$. 

% The DDPM framework comprises two primary phases: 
The method for pre-training our base DDPM includes two stages:
the\textbf{ forward diffusion process} and the\textbf{ reverse denoising process}~\citep{ho2020denoising}. In the forward process, a Markov chain is constructed to progressively add noise to the original data sample $x_0 \sim p_{\text{data}}$ by $x_t=\sqrt{1-\beta_t}x_{t-1}+\sqrt{\beta_t}\boldsymbol{\epsilon}_t$ to get $x_1,x_2,\cdots,x_T$. The mathematical formulation of the forward diffusion process is given by:
\begin{equation}
q(x_1,\cdots,x_T|x_0)=\prod_{t=1}^{T}q(x_t|x_{t-1}),
\end{equation}
where $t$ denotes as diffusion step, $q(x_t|x_{t-1})= \mathcal{N}(x_t;\sqrt{1-\beta_t}x_{t-1}, \beta_tI)$, and $\beta_t$ are variance schedule.  
According to the additivity property of Gaussian distributions, the distribution of $x_t$ in the forward diffusion process at step $t$ is formulated by $x_t=\sqrt{\bar{\alpha}}x_0+\sqrt{1-\bar{\alpha}}\boldsymbol{\epsilon}_t $, where $\epsilon\sim \mathcal{N}(0,\textbf{I})$,$\alpha_t=1-\beta_t$ and $\bar\alpha=\prod_{i=1}^{t}\alpha_i$. Therefore, $q(x_t|x_0)=\mathcal{N}(x_{t};\sqrt{\bar{\alpha}}x_0,(1-\bar{\alpha})I)$. $\bar\alpha$ goes to 0 when $t$ goes to large T, and eventually  $x_T\sim\mathcal{N}(0,1)$. 

In the reverse process, we sample reverse steps $q(x_{t-1}|x_t)$ until reaching $x_0$, which can also be a Markov Process defined as:
\begin{equation}
p_\theta(x_0,\cdots,x_{T-1}|x_T)=\prod_{t=1}^{T}p_\theta(x_{t-1}|x_t).
\end{equation}
To make the reverse progress learnable, ~\citet{sohl2015deep} proposed to train a neural network to predict a mean $\mu_\theta$ and a diagonal covariance matrix $\Sigma_\theta$:
\begin{align}
q(x_{t-1}|x_t, x_0) = \mathcal{N}(x_{t-1};\tilde\mu(x_t, x_0), \tilde\beta_tI),\\
\tilde\mu(x_t, x_0)=\frac{\sqrt{\bar\alpha_{t-1}}\beta_t}{1-\bar\alpha_t}x_0+\frac{\sqrt{\alpha_t}(1-\bar\alpha_{t-1})}{1-\bar\alpha_t}x_t.
\end{align}
The ReconMOST effectively learns the latent horizontal and vertical distribution patterns of ocean temperature fields and their inter-layer interactions. Random samples generated from the model naturally exhibit spatial structures consistent with fundamental oceanographic principles.

\subsection{Observation Guided Reconstruction}

We perform conditional sampling reconstruction based on the pretrained DDPM model, using a set of known observations, such as buoy data, as constraints. Specifically, the sampling process is guided by gradients derived from these observations, enabling the reconstruction of accurate multi-layer global sea temperature fields from sparse observation points. The reconstruction procedure is detailed in Algo.~\ref{alg:guided-sample}.

\subsubsection{Gradient Guidance}

In this conditional diffusion setting, the condition is defined by the observed sea temperature data $x_0$, represented as a sparsely masked $y$ where only a few observation points remain, consistent with buoy-type data. 
The main objective is to reconstruct an accurate global sea temperature $x_0$ conditional on sparse observations $y$ and intermediate sampling result $x_t$.

% According to Bayes' theorem:
% \begin{align}
%     p(x_t|x_{t+1}, y) & =\frac{p(x_t,x_{t+1},y)}{p(x_{t+1},y)}=\frac{p(x_t|x_{t+1})p(x_{t+1})p(y|x_t,x_{t+1})}{p(y|x_{t+1})p(x_{t+1})}\\ &=\frac{p(x_t|x_{t+1})p(y|x_t,x_{t+1})}{p(y|x_{t+1})}=\frac{p(x_t|x_{t+1})p(y|x_t)}{p(y|x_{t+1})}\\ &\propto p(x_t|x_{t+1})p(y|x_t),
% \end{align}

% where the penultimate step is derived from :
% \begin{equation}
% p(y|x_t,x_{t+1})=p(x_{t+1}|x_t,y)\frac{p(y|x_t)}{p(x_{t+1}|x_t)}=p(x_{t+1}|x_t)\frac{p(y|x_t)}{p(x_{t+1}|x_t)}=p(y|x_t).
% \end{equation}

~\citet{sohl2015deep} has proven that:
\begin{equation}
\log p_\theta(x_{t-1}|x_t,y)=\log \left(p_\theta(x_{t-1}|x_t)p(y|x_t)\right)+C_1\approx \log p(z)+C_2,
\end{equation}
where $z\sim \mathcal{N}(\mu(x_t,t)+\Sigma g,\Sigma)$ and $g=\nabla_{x_t}\log p(y|x_t)|_{x_t=\mu(x_t,t)}$, $\Sigma=\Sigma_\theta(x_t)$. ~\citet{dhariwal2021diffusion}, and ~\citet{ho2022classifier} have provided concrete implementations.

For $p(y|x_t)$, we make an intuitive assumption that the probability of generating observations sea temperature data $y$ from $x_t$ is equivalent to that from the reconstructed $\hat x_0$ obtained via directly reverse diffusion from $x_t$. Accordingly, this conditional probability can be evaluated by the difference between reconstruction sea temperature $\hat{x}_0$ and observation sea temperature $y$:
\begin{align}
&p(y|x_t)\approx\exp{(-s\cdot L(\hat x_0(x_t),y))},\\
&\nabla_{x_t}\log p(y|x_t)\approx -s\cdot\nabla_{x_t}L(\hat x_0(x_t),y).
\end{align}
At each reconstructing(denoising) step from $x_t$ to $x_{t-1}$, we steer the distribution of $x_{t-1}$ towards the actual sea temperature distribution associated with the observed values y. This is achieved by computing the discrepancy between the reconstructed $\hat{x}_0$ from $x_t$ and the observations y, and using the resulting gradient to adjust the reconstructing direction of $x_t$~\citep{fei2023generative}.
% 新加的，还是把实现中对xt求梯度简化成x0放到这里来了，不然感觉实验的时候再提容易混淆，伪代码也跟着修改了
In this work, we approximate the gradient with respect to $x_t$ by that of $x_0$, namely $g=\nabla_{x_0}\log p(y|x_t)|_{x_t=\mu(x_t,t)}$, which preserves the effectiveness of the heuristic formulation for $p(y|x_t)$ and achieves satisfactory reconstruction accuracy.
By incorporating this gradient-based guidance into the sampling process, the method not only improves reconstruction accuracy at the observation points but also leverages the physical consistency embedded in the pretrained model to produce plausible and coherent reconstructions in unobserved regions.

% \begin{algorithm}[t]
% \renewcommand{\baselinestretch}{1.2}\selectfont % 设置行距1.2倍
% \caption{\textbf{Observation Guided Reconstruction}}
% \label{alg:guided-sample}
% \begin{algorithmic}[1]  % [1] 表示行号起始
%     \State \textbf{Input:} Sparse observation data $y$, gradient scale $s$, distance measure $\mathcal{L}$, Gaussian Kernal $\mu,\sigma$.
%     \State \textbf{Output:} Output image $x_0$ conditioned on $y$
%     \State Sample $x_T$ from $\mathcal{N}(0, \mathbf{I})$
%     \For{$t$ from $T$ to $1$}
%         \State $\hat x_0 = \frac{x_t - \sqrt{1-\alpha_t} \epsilon_\theta(x_t, t)}{\sqrt{\alpha_t}}$
%         \State $g = -s\cdot\nabla_{x_0}L(\hat x_0,y)$
%         \State $g=\text{SoftExtension}(g,\mu,\sigma)$
%         \State Sample $x_{t-1} \sim \mathcal{N}\left(\mu + \Sigma g, \Sigma \right)$
%     \EndFor
%     \State \Return $x_0$
% \end{algorithmic}
% \end{algorithm}

\subsubsection{Extension of Observation Guidance}
\begin{wrapfigure}{r}{0.45\textwidth}
  \centering
  \vspace{-26pt}
    \includegraphics[width=0.4\textwidth]{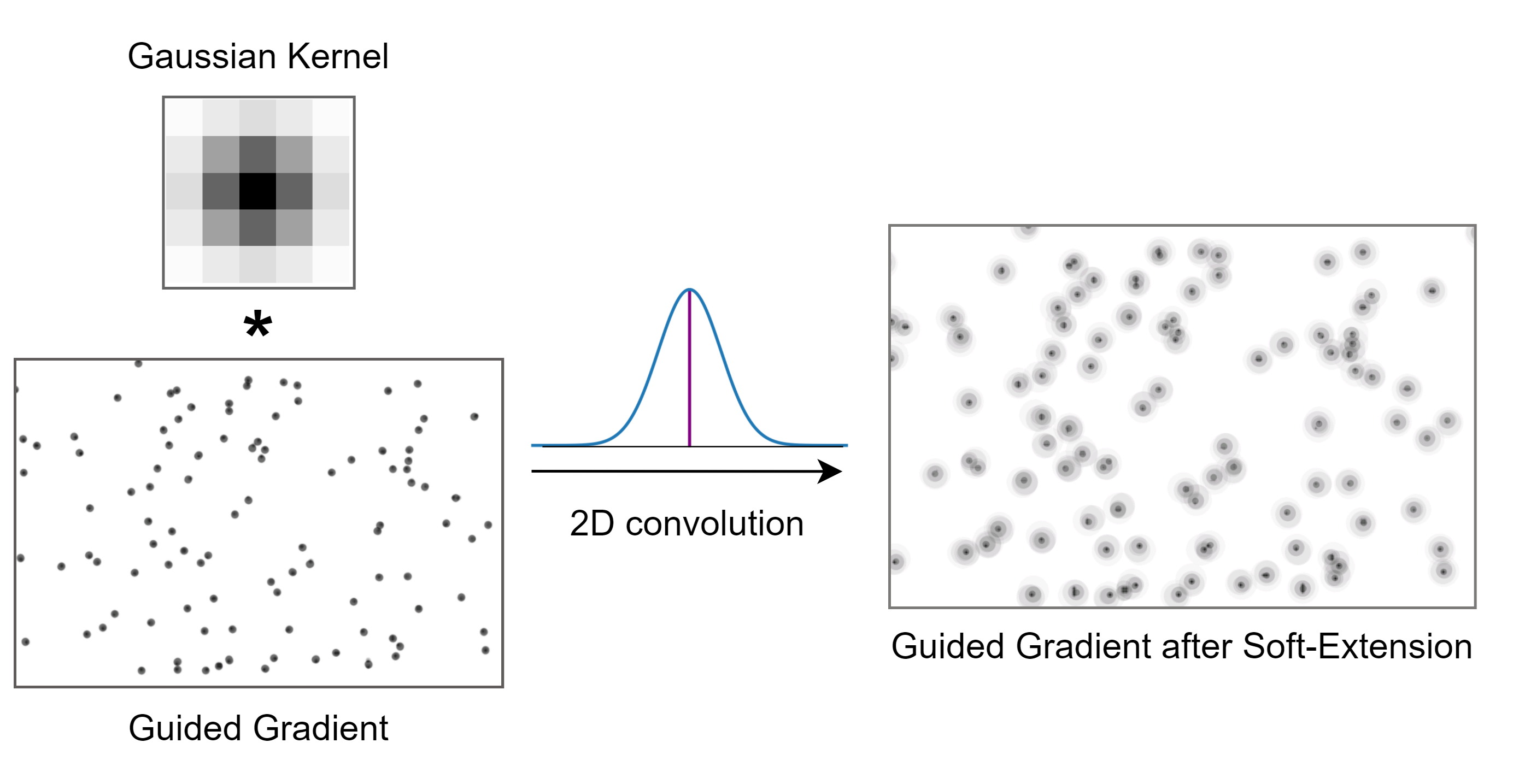}
  \caption{Demonstration of Soft-Extension}
  \label{fig:soft-extension}
  \vspace{-12pt}
\end{wrapfigure}
Given the extreme sparsity of current buoy observations, directly applying observation-derived gradients results in insufficient constraints for unobserved regions. Inspired by Tobler's first law of geography, which states that all spatial variables are related, with nearby values being more strongly correlated~\citep{tobler1970computer,huang2024diffda}, we introduce a Soft-Extension strategy. Without altering the sparsity pattern of observation points, this method spatially diffuses the gradients $g$ of observed points to their neighbors via a Gaussian kernel as Fig.~\ref{fig:soft-extension}. Since sea temperature fields typically exhibit continuous and smooth spatial variations, nearby points are expected to have similar values. The Soft-Extension effectively enhances local reconstruction accuracy while preserving the sparse observational constraints.

\textbf{Before Extension:} Only observed points carry gradients, leaving surrounding regions unconstrained.
        
\textbf{After Extension:} Gradients at observed points are spatially diffused to neighboring regions with a Gaussian weighting, enabling nearby points to acquire indirect constraints and improving local reconstruction quality.

% \begin{figure}[ht]
%     \centering
%     \begin{minipage}[t]{\textwidth}
%     \begin{minipage}[t]{0.45\textwidth}
%         % \small  % 可以调字体大小
%         \vspace{0pt}
%         \textbf{Before Extension} Only observed points carry gradients, leaving surrounding regions unconstrained.
        
%         \textbf{After Extension} Gradients at observed points are spatially diffused to neighboring regions with a Gaussian weighting, enabling nearby points to acquire indirect constraints and improving local reconstruction quality.
%     \end{minipage}
%     \hfill
%     \begin{minipage}[t]{0.5\textwidth}
%         \vspace{0pt}
%         \centering
%         \includegraphics[width=\linewidth]{Figures/softmask.png}
%         \caption{demonstration of soft-extension}
%         \label{fig:right_half}
%     \end{minipage}
%     \end{minipage}
% \end{figure}

We do not perform additional extensions in the vertical dimension for two main reasons. First, the dataset exhibits heterogeneous depth intervals and representational scales across different vertical layers, which may compromise spatial consistency if directly extended. Second, the vertical distribution of sea temperature typically follows a relatively stable pattern compared to the horizontal direction, making additional model extensions in the vertical dimension unnecessary. Thereby, we avoid redundant complexity and improving computational efficiency.

\section{Experiments and Analysis}
\label{sec:experiments}

\begin{figure}[t]
    \centering
    \includegraphics[width=1.0\columnwidth]{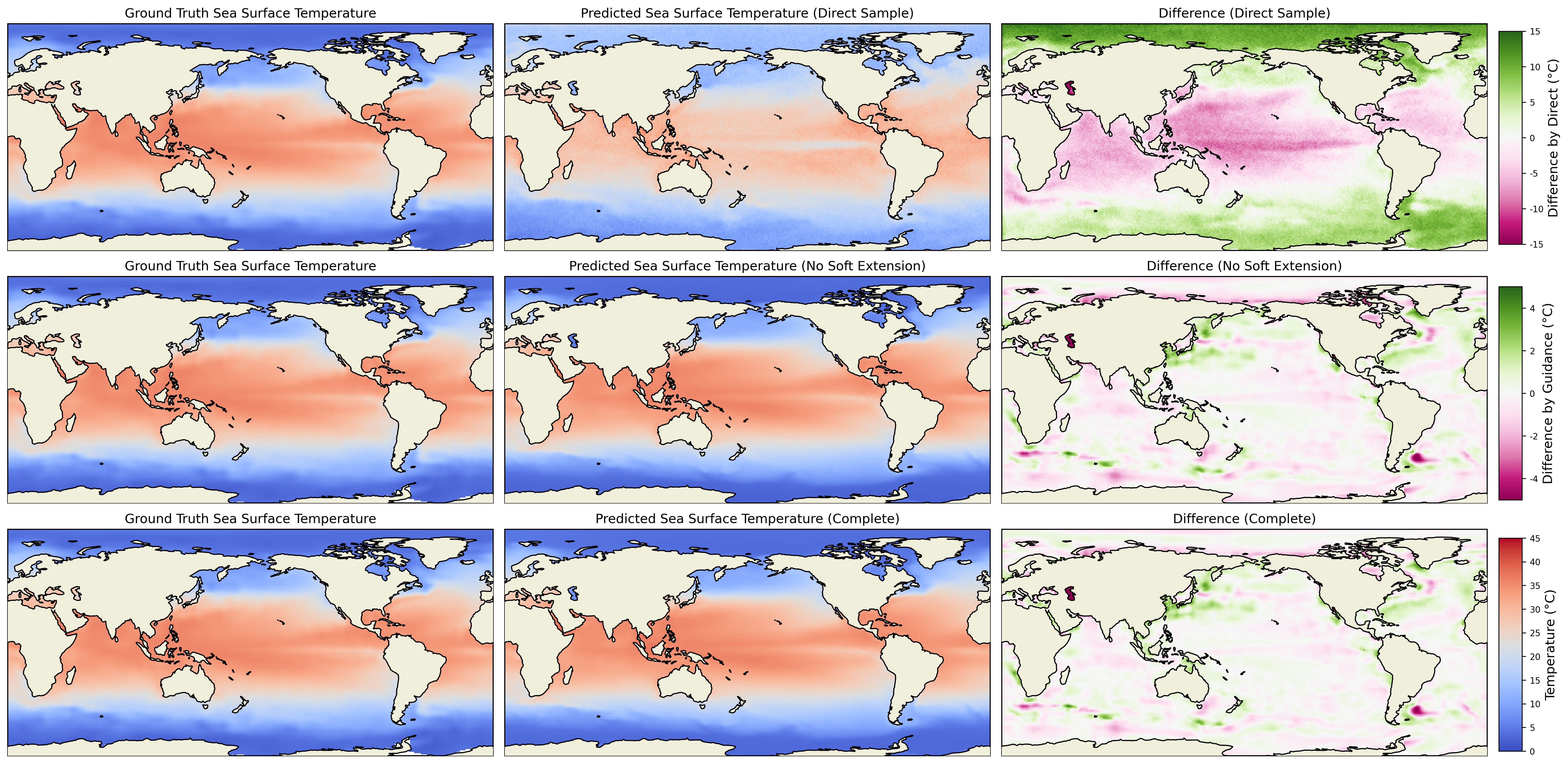}
    \vspace{-0.2cm}
    \caption{Comparison of SST reconstruction of EN4 dataset in ground truth, reconstruction results, and difference: for each role, the figure shows ground truth, reconstruction result and difference of SST between ground truth and reconstruction result with no guidance, guidance with no soft extension, and complete guidance.}
    \label{fig:result}
\vspace{-0.4cm}
\end{figure}

\subsection{Datasets}

% Since the monthly ocean data is scarce, to increase the number of training data and reduce the systematic biases among dynamical models, historical simulations from 7 models (table xx) that participated in the Coupled Model Intercomparison Project phase 6 (CMIP6) \citep{eyring2016overview} are used for training. 
\textbf{CMIP6.}\quad
Due to the scarcity of monthly ocean data, we utilize historical simulations from seven models that participated in the Coupled Model Intercomparison Project Phase 6 (CMIP6)~\citep{eyring2016overview} to increase the amount of training data and also reduce systematic biases among dynamical models.
CMIP6 comprises a suite of comprehensive Earth system models developed by leading climate research centers to simulate past, present, and future climate conditions under standardized experimental protocols. 
Here, historical simulations from these Earth system models can help our model obtain more robust predictions that capture the interconnected dynamics across ocean basins, thus enhancing model generalization and predictive capability. \\
\textbf{EN4}\quad is a global ocean temperature and salinity dataset developed by the UK Met Office~\citep{EN4_2013}, providing quality-controlled subsurface ocean profile observations from various sources, including Argo floats~\citep{roemmich2009argo}, expendable bathythermographs (XBTs)~\citep{goni2019XBT}, and ship-based measurements. 
Available from 1900 to the present, EN4 offers monthly objective analyses of temperature and salinity at standard depth levels, making it a valuable resource for climate studies and ocean model validation. 
In this study, we utilize upper ocean temperature data from EN4 during the 21st century (i.e., after 2000) as ground truth to guide and evaluate our proposed method for complete ocean temperature reconstruction.

\subsection{Experimental Setups}
The DDPM model is configured as follows: `in\_channels' is 42, corresponding to 42 ocean temperature layers. The Unet backbone adopts channel-mult (1, 2, 2, 2, 4, 4), progressively increasing feature channels to capture multi-scale spatial patterns. The base number of channels is set to 128, with 3 residual blocks per layer. The number of diffusion steps is 1000, with a linear noise schedule. 
In the \textbf{training phase}, the learning rate is 1e-4, and the batch size is 4. 
We trained each model for 200,000 steps, requiring approximately 24 hours on a single A100 GPU.
In the \textbf{inference phase}, we independently reconstructed each observation sparsity month four times. Observation points are randomly sampled from the global sea temperature field according to the specified `guided\_rate', with consistent ratios across all layers but varied locations in each trial. The final performance is evaluated by averaging the MSE over four reconstruction results with ground truth temperature. The default setting of `guided\_rate' is 0.075, which means selecting 7.5\% of grid points as known observations, matching real-world buoy distribution densities. During reconstruction, the Euclidean distance between $\hat{x}_0(x_t)$ and observation $y$ is used as the guided distance function to compute the gradient-based guidance term. Additionally, $\mathbf{\Sigma}$ can optionally be set to 0 (controlling noise influence), while the \textbf{guidance strength $\textit{\textbf{s}}$} and the \textbf{Gaussian kernel size for soft-extension} are treated as hyperparameters, alterable according to observation sparsity and spatial distribution characteristics.

\subsection{Effectiveness Across Various Data Modalities}
\label{sec:experiments_effectiveness}
% or Effectiveness Assessment
% Vaidation of Effectiveness
% This experiment aims to validate the effectiveness of the proposed method, specifically whether the pretrained model can maintain physical consistency and accurately reconstruct the global multi-layer sea temperature fields with sparse observational guidance. 
% We selected seven CMIP6 models with the largest data volume, each with 3 or 5 ensemble members spanning 165 years (1980 samples). 
To validate the effectiveness of the proposed method, we selected seven modes in CMIP6 with large data volume to conduct experiments.  
Our ReconMOST works across all seven modes.
% and the best accuracy of the result achieved an amazing 0.063 (Table~\ref{tab:table1}).
Specifically, for each model with 3 or 5 initializations spanning 165 years (1980 samples), independent training and testing were conducted by reserving the last initialization as the test set and using the others for training. 
Only data from 2000 to 2014 (180 samples) were used for testing, as the number of buoy observations before 2000 was too limited for practical application.
As described in the settings, each case was reconstructed four times and computed the mean MSE. 
The results in Table~\ref{tab:table1} demonstrate that the pretrained model effectively captures the implicit physical patterns of sea temperature distributions when the training and testing models are consistent. 
The best-performing model achieved the reconstruction with the lowest MSE of 0.063.
% The best-performing model achieved an accuracy of reconstruction of an amazing 0.063.

% \captionsetup{justification=centering}
\begin{table}[t]
    % \captionsetup[table]{justification=centering}  
    \caption{Guided reconstruction result of CMIP6: train and validation datasets are from the same mode, and the last three columns are respectively average MSE of upper ten levels on guidance points (MSE-g), reconstruction points (MSE-r) and all complete global points (total MSE).}
    \vspace{+0.2cm}
    \centering
    % 定义表格列格式：
    % | 表示竖线，l 表示左对齐，c 表示居中，r 表示右对齐
    % 竖线放在 mode 和 s、softmask 和 loss1 之间
    % \resizebox{\textwidth}{!}{
    \begin{tabular}{l c c c c c c c}
        \toprule
        % 表格标题行
        \textbf{mode} & \textbf{s} & \textbf{sigma} & \textbf{guided rate(\#)} & \textbf{soft-ext size} & \textbf{MSE-g} & \textbf{MSE-r} & \textbf{total MSE} \\
        \midrule
        % 数据行（示例数据）
        FGOALS-f3-L  & 4 & 0 & 7.5\%(3188) & 5 & 0.046& 0.607& 0.565\\
        CanESM5  & 4 & 0 & 7.5\%(3188)  & 5 & 0.043 & 0.559 & 0.521 \\
        IPSL-CM6A-LR  & 4 & 0 & 7.5\%(3188) & 5 & 0.044 & 0.585 & 0.544 \\
        FGOALS-g3  & 4 & 0 & 7.5\%(3188) & 5 & 0.045 & 0.585 & 0.545 \\
        BCC-CSM2-MR  & 4 & 0 & 7.5\%(3188) & 5 & 0.033 & 0.428 & 0.398 \\
        FIO-ESM-2-0  & 4 & 0 & 7.5\%(3188) & 5 & 0.005 & 0.068 & 0.063 \\
        MRI-ESM2-0  & 4 & 0 & 7.5\%(3188) & 5 & 0.025 & 0.330 & 0.308 \\
        \bottomrule
    \end{tabular}
    % }
    \label{tab:table1}
\vspace{-0.4cm}
\end{table}

\begin{table}[ht]
    \centering
    \caption{Guided reconstruction result of EN4 analysis: train dataset is CMIP6 historical numerical simulation profile, whose total MSE of guided reconstruction is top k least on EN4 analysis, and test dataset is EN4 analysis. The last three columns are respectively average MSE of upper ten levels on guidance points (MSE-g), reconstruction points (MSE-r) and all complete global points (total MSE).}
    \vspace{+0.2cm}
    % 定义表格列格式：
    % | 表示竖线，l 表示左对齐，c 表示居中，r 表示右对齐
    % 竖线放在 mode 和 s、softmask 和 loss1 之间
    % \resizebox{0.9\textwidth}{!}{
    \begin{tabular}{l c c c c c c c}
        \toprule
        % 表格标题行
        \textbf{mode} & \textbf{s} & \textbf{sigma} & \textbf{guided rate(\#)} & \textbf{soft-ext size} & \textbf{MSE-g} & \textbf{MSE-r} & \textbf{total MSE} \\
        \midrule
        % 数据行（示例数据）
        Top1  & 4 & 0 & 7.5\%(3188) & 5 & 0.055 & 0.737 & 0.686\\
        Top3  & 4 & 0 & 7.5\%(3188)  & 5 & 0.091 & 1.229 & 1.144 \\
        Top5  & 4 & 0 & 7.5\%(3188) & 5 & 0.051 & 0.696 & 0.648\\
        Top7  & 4 & 0 & 7.5\%(3188) & 5 & 0.053 &  0.711 & 0.662\\
        \bottomrule
    \end{tabular}
    % }
    \label{tab:table2}
\vspace{-0.4cm}
\end{table}

\subsection{Generalization Capabilities Across Diverse Scenarios}
\label{sec:experiments_generalization}
% or Generalization Assessment
% Vaidation of Generalization
This experiment aims to evaluate the generalization capability of our ReconMOST. 
We investigate whether a pretrained model can achieve high-precision multi-layer sea temperature reconstruction on datasets derived from unseen simulation modes. 
This setting closely reflects real-world application scenarios, where temporal, climatic, and noise variations pose significant challenges to model generalization. 
We applied the seven models pretrained during the effectiveness verification phase to the EN4 analysis dataset, using global average MSE as the evaluation metric, where a lower MSE indicates stronger generalization. 
Additionally, we selected the Top1, Top3, and Top5 modes (ranked by MSE performance), along with the complete seven-mode dataset, to retrain DDPM models.  
This strategy balances generalization capacity with training data volume to capture more diverse distribution patterns. 
The results in Table~\ref{tab:table2} show that the Top5 model achieves the best performance of 0.648, while the Top3 performs worst, likely due to increased distributional variance arising from a smaller mixture of modes. 
Meanwhile, Top1 model also performed well, with only 0.038 difference from the best model, which means that we can achieve good generalization performance by learning from just one single best mode. 
So as long as the pre-trained analytical data physicality is good enough, one mode is sufficient.
Moreover, we visualize our reconstruction results as shown in Fig.~\ref{fig:result}, where the first row is the unconditional generation result using the pretrained model, and the other rows are conditional generation results guided by observations. 
Although there is a noticeable deviation between the unconditional generation result and the ground truth, the spatial patterns of temperature variation remain physically consistent.
% Although there is a noticeable deviation between the direct result and the ground truth, the spatial patterns of temperature variation remain physically consistent.
% These findings suggest that a trade-off between generalization ability and dataset diversity is necessary for optimal reconstruction performance.

\subsection{Scalability Under Increasing Observation Density}
% Scalability Assessment
% Vaidation of Scalability
To evaluate the scalability of the proposed method under varying observational guidance ratios, we conducted reconstruction experiments with different proportions of observational points, considering potential future increases in buoy coverage.
The results in Table~\ref{tab:table3} indicate a consistent and significant improvement in the accuracy of reconstruction as the guidance ratio increases. 
This observation is intuitive, as a higher density of observational points provides richer spatial information for reconstructing the temperature fields, thereby enhancing model performance.
We hope that in the future, more buoy observations combined with our method will contribute to a better understanding of the ocean.

\begin{table}[t]
    \centering
    \caption{Guided reconstruction result of EN4 analysis with difference guided rate (number of guided points): use best model Top5 in Table~\ref{tab:table2} and other parameters s=4, Sigma=0, soft-ext size is 5.The last three columns are, respectively, average MSE of upper ten levels on guidance points (MSE-g), reconstruction points (MSE-r), and all complete global points (total MSE).}
    \vspace{+0.2cm}
    % \caption{Guided Reconstruction Result of EN4 Analysis with Difference Guided Rate(number of guided points): Use best model Top5 in table~\ref{tab:table2} and other parameters s=4, Sigma=0, soft-ext size is 5.The last three columns are respectively average MSE of upper ten levels on guidance points(MSE-g), reconstruction points(MSE-r) and all complete global points(total MSE).}
    % 定义表格列格式：
    % | 表示竖线，l 表示左对齐，c 表示居中，r 表示右对齐
    % 竖线放在 mode 和 s、softmask 和 loss1 之间
    % \resizebox{0.8\textwidth}{!}{
    \begin{tabular}{c c c c c}
        \toprule
        % 表格标题行
        \small\textbf{guided rate} & \small\textbf{\# guided points(round)}& \small\textbf{MSE-g} & \small\textbf{MSE-r} & \small\textbf{total MSE} \\
        \midrule
        % 数据行（示例数据）
        7.5\%&3188 & 0.051 & 0.696 & 0.648 \\
        \midrule
        10\%&4251 & 0.068 & 0.672 & 0.612\\
        20\%&8502 & 0.132 & 0.563 & 0.476\\
        30\%&12754 & 0.151 & 0.370 & 0.305\\
        40\%&17005 & 0.165 & 0.259 & 0.222 \\
        \bottomrule
    \end{tabular}
    % }
    \label{tab:table3}
\vspace{-0.4cm}
\end{table}
% 缺一个wo，不加任何改进的

\subsection{Ablation Studies}
% We performed ablation experiments of the hyperparameters in the proposed method.

\textbf{Guidance Strength $\textit{\textbf{s}}$.}\quad
In the reconstruction process, $s$ controls the constraint strength imposed by the distance function on the observational guidance points. Theoretically, a smaller $s$ results in a higher heuristic probability $p(y|x_t)$ for a given distance, allowing the reconstruction to tolerate greater deviations from the observed values while still achieving satisfactory global performance. To investigate the impact of $s$, we conducted an ablation study using the optimal Top5 model under varying integer values of $s$. The results in Table~\ref{tab:table4} reveal that the accuracy of reconstruction initially improves and subsequently declines as $s$ increases. This suggests that excessively strict constraints enforcing exact matching with observations may impair overall reconstruction performance. Therefore, selecting an appropriate $s$ requires balancing constraint intensity and global accuracy of reconstruction, potentially in conjunction with other hyperparameters and the spatial distribution of observation points.

\begin{table}[t]
    \centering
    \caption{MSE result of ablation on \textit{sigma}, \textit{s}, \textit{softmask scale}. The last three columns are respectively average MSE of upper ten levels on guidance points (MSE-g), reconstruction points (MSE-r) and all complete global points (total MSE).}
    \vspace{+0.2cm}
    % \small
    % 定义表格列格式：
    % | 表示竖线，l 表示左对齐，c 表示居中，r 表示右对齐
    % 竖线放在 mode 和 s、softmask 和 loss1 之间
    % \resizebox{\textwidth}{!}{
    \begin{tabular}{l c c c c c c c}
        \toprule
        % 表格标题行
        \textbf{Config} & \textbf{s} & \textbf{sigma} & \textbf{guided rate(\#)} & \textbf{soft-ext size} & \textbf{MSE-g} & \textbf{MSE-r} & \textbf{total MSE} \\
        \midrule
        % 数据行（示例数据）
        Top5  & 4 & 0 & 7.5\%(3188) & 5 & 0.051 & 0.696 & 0.648 \\
        \midrule
        S-3  & 3 & 0 & 7.5\%(3188) & 5 & 0.055 & 0.739 & 0.688 \\
        S-5  & 5 & 0 & 7.5\%(3188) & 5 & \textbf{0.049} & \textbf{0.680} & \textbf{0.633}\\
        S-6  & 6 & 0 & 7.5\%(3188) & 5 & 0.049 & 0.689 & 0.641 \\
        \midrule
        EXT-3  & 4 & 0 & 7.5\%(3188) & 3 & 0.055 & 0.754 & 0.702\\
        EXT-7  & 4 & 0 & 7.5\%(3188) & 7 & 0.052 & 0.717 & 0.667\\
        EXT-9  & 4 & 0 & 7.5\%(3188) & 9 & 0.051 & 0.701 & 0.653\\
        \midrule
        W.O Sigma=0 & 4 & DDPM & 7.5\%(3188) & 5 & 0.052 & 0.721 & 0.671\\
        W.O. Extension & 4 & 0 & 7.5\%(3188) & 0 & 0.112 & 1.479 & 1.376 \\
        W.O. Sigma \& Ext & 4 & DDPM & 7.5\%(3188) & 0 & 0.113 & 1.487 & 1.384\\
        \bottomrule
    \end{tabular}
    % }
    \label{tab:table4}
\vspace{-0.4cm}
\end{table}
% 缺一个wo，不加任何改进的

\textbf{W.O. \textit{Sigma}.}\quad
Experimental results indicate that during the gradient-guided sampling process, retaining the original diffusion variance $\Sigma$ when computing the guiding term g for updating $\mu$, while setting the diffusion variance $\Sigma$ to zero in the sampling step from $x_t$ to $x_{t-1}$ with new $\mu$, significantly improves the accuracy of reconstruction. This configuration avoids unnecessary noise accumulation in the diffusion process, stabilizing the sampling trajectory and enhancing reconstruction performance under sparse observational guidance (Table~\ref{tab:table4}).

\textbf{Extension Kernel Size}\quad
In the soft-extension module, the Gaussian kernel size determines the horizontal influence range of temperature observations. 
A larger kernel implies a stronger guiding effect of individual observation points on their surrounding areas. 
Experimental results from Table~\ref{tab:table4} indicate that increasing the kernel size consistently improves the accuracy of reconstruction. 
Without applying the extension operation, the reconstruction performance becomes significantly worse. 
This can be attributed to the highly sparse distribution of observations, where local guidance alone is insufficient, and spatial diffusion of observation influence effectively enhances reconstruction performance.
The second row in Fig.~\ref{fig:result} illustrates the reconstruction results guided by sparse observations without applying soft-extension. Compared to the third row, which uses the complete ReconMOST framework, it is evident that ReconMOST achieves superior reconstruction accuracy in both open ocean and coastal regions. 
This highlights the critical role of soft-extension in enhancing reconstruction performance.
These findings further validate the effectiveness of our ReconMOST.

\section{Conclusion and Discussion}
\label{sec:conclusion}
This paper proposes ReconMOST, a guided diffusion-based framework for ocean reconstruction tasks, achieving for the first time high-accuracy global multi-layer sea temperature field reconstruction. The method demonstrates excellent performance even under extremely sparse observation scenarios, with up to 92.5\% of missing data, effectively addressing the limitations of the current global buoy observation network.

Despite its promising performance, several aspects of our ReconMOST might be further improved. 
First, we think the relationship between the guidance strength $s$, the soft-extension kernel size, and both the number and spatial distribution of observation data remains underexplored. Establishing a functional relationship or adaptive adjustment strategy is expected to enhance reconstruction accuracy. 
% Second, this study only focuses on sparsely distributed observation scenarios, without evaluating the reconstruction quality under densely clustered observational conditions.
Second, the current framework is evaluated to static three-dimensional spatial reconstructions without incorporating temporal dynamics. Future work will extend our framework to spatiotemporal reconstructions.

Most importantly, ReconMOST provides a novel methodology and perspective for ocean reconstruction problems, with excellent generalization and scalability. 
It holds the promising potential for extension to other oceanic variables and diverse observation conditions, contributing to improved ocean monitoring, climate prediction, and sustainable marine resource development.

% Here is an example usage of the two main commands (\verb+citet+ and \verb+citep+): Some people thought a thing \citepp{kour2014real, hadash2018estimate} but other people thought something else \citepp{kour2014fast}. Many people have speculated that if we knew exactly why \citept{kour2014fast} thought this\dots

% The documentation for \verb+booktabs+ (`Publication quality tables in LaTeX') is available from:
% \begin{center}
% 	\url{https://www.ctan.org/pkg/booktabs}
% \end{center}

% Citations use \verb+natbib+. The documentation may be found at
% \begin{center}
% \url{http://mirrors.ctan.org/macros/latex/contrib/natbib/natnotes.pdf}
% \end{center}

% \begin{table}
% 	\caption{Sample table title}
% 	\centering
% 	\begin{tabular}{lll}
% 		\toprule
% 		\multicolumn{2}{c}{Part}                   \\
% 		\cmidrule(r){1-2}
% 		Name     & Description     & Size ($\mu$m) \\
% 		\midrule
% 		Dendrite & Input terminal  & $\sim$100     \\
% 		Axon     & Output terminal & $\sim$10      \\
% 		Soma     & Cell body       & up to $10^6$  \\
% 		\bottomrule
% 	\end{tabular}
% 	\label{tab:table}
% \end{table}

\bibliographystyle{unsrtnat}
\bibliography{references}  
%%% Uncomment this line and comment out the ``thebibliography'' section below to use the external .bib file (using bibtex) .





% \end{thebibliography}

\newpage
\appendix
\section{Derivation of Denoising Diffusion Probabilistic Models}
% Technical appendices with additional results, figures, graphs and proofs may be submitted with the paper submission before the full submission deadline (see above), or as a separate PDF in the ZIP file below before the supplementary material deadline. There is no page limit for the technical appendices.

Consider a data distribution $q(x_0)$ over continuous data $x_0\in\mathbb{R}^d$. DDPM introduces a latent variable sequence $\{x_1, \dots, x_T\}$ by defining a Markov chain that gradually adds Gaussian noise to the data~\citep{ho2020denoising}:
\begin{align}
 q(x_{1:T}|x_0)& = \prod_{t=1}^{T} q(x_t|x_{t-1})\\
 q(x_t|x_{t-1})& = \mathcal{N}(x_t;\sqrt{1-\beta_t} \, x_{t-1}, \beta_t \mathbf{I}),
\end{align}
where $$\{\beta_t\}_{t=1}^{T}$$ is a predefined variance schedule satisfying $0 < \beta_t < 1$.
Using the properties of Gaussians, the marginal distribution at any time t can be derived as:
\begin{equation}
q(x_t|x_0) = \mathcal{N}(x_t; \sqrt{\bar{\alpha}_t} \, x_0, (1-\bar{\alpha}_t)\mathbf{I}).
\end{equation}
where $\bar{\alpha}_t = \prod_{s=1}^{t} (1-\beta_s)$.
The generative process aims to learn the reverse Markov chain~\citep{ho2020denoising}:
\begin{equation}
    p_\theta(x_{0:T}) = p(x_T) \prod_{t=1}^{T} p_\theta(x_{t-1}|x_t)
\end{equation}
with prior $p(x_T) = \mathcal{N}(x_T; 0, \mathbf{I})$ and parameterized reverse transitions:
\begin{equation}
p_\theta(x_{t-1}|x_t) = \mathcal{N}(x_{t-1}; \mu_\theta(x_t, t), \Sigma_\theta(x_t, t)).
\end{equation}
By minimizing the variational bound on the negative log-likelihood:
\begin{equation}
\mathbb{E}_q \left[-\log p_\theta(x_0 | x_1) + \sum_{t=2}^{T} D_{KL}\left(q(x_{t-1}|x_t, x_0) \| p_\theta(x_{t-1}|x_t)\right) + D_{KL}\left(q(x_T|x_0) \| p(x_T)\right)\right],
\end{equation}
the model can be trained efficiently. It is further simplified by noting that the true posterior $q(x_{t-1}|x_t, x_0)$ is Gaussian with known closed-form expressions for its mean and variance:
\begin{equation}
q(x_{t-1}|x_t, x_0) = \mathcal{N}(x_{t-1}; \tilde{\mu}(x_t, x_0), \tilde{\beta}_t \mathbf{I}),
\end{equation}
where
\begin{equation}
\tilde{\mu}(x_t, x_0) = \frac{\sqrt{\bar{\alpha}_{t-1}} \, \beta_t}{1-\bar{\alpha}_t} x_0 + \frac{\sqrt{1-\beta_t} (1-\bar{\alpha}_{t-1})}{1-\bar{\alpha}_t} x_t,
\end{equation}
and $\tilde{\beta}_t = \frac{1-\bar{\alpha}_{t-1}}{1-\bar{\alpha}_t}\beta_t$.
However, in practice, we do not have access to the true $x_0$ during sampling, we use a rearranged forward process formula to express $x_0$ as:
\begin{equation}
\hat{\mathbf{x}}_0 = \frac{\mathbf{x}_t - \sqrt{1 - \bar{\alpha}_t} \epsilon_\theta}.{\sqrt{\bar{\alpha}_t}}   
\end{equation}
By substituting this estimated $x_0$ into the expression for $\tilde{\mu}_t$, we obtain an approximate mean for the reverse process transition:
\begin{equation}
\mu_\theta({x}_t, t) = \frac{\sqrt{\bar{\alpha}_{t-1}} \beta_t}{1 - \bar{\alpha}_t} \hat{{x}}_0 + \frac{\sqrt{\alpha_t} (1 - \bar{\alpha}_{t-1})}{1 - \bar{\alpha}_t}{x}_t.
\end{equation}
This formulation allows us to iteratively sample $x_{t-1}$ from $x_t$ using:
\begin{equation}
p_\theta({x}_{t-1}|{x}_t) = \mathcal{N}\left(\mathbf{x}_{t-1}; \boldsymbol{\mu}_\theta(\mathbf{x}_t, t), {\Sigma}_\theta(t) \right),
\end{equation}

until reaching ${x}_0$ at the final step.
In practice, the network is trained to predict the noise term $\epsilon$ using a U-Net network whose loss function is as follows:
\begin{equation}
L:=E_{t\sim[1,T],x_0 \sim q(x_0),\epsilon\sim N (0,\mathbf{I})}[\|\epsilon-\epsilon_\theta(x_t,t)\|^2].
\end{equation}
Note that L does not provide any learning signal for $\Sigma_\theta(xt, t)$. ~\citet{ho2020denoising} find that instead of learning $\Sigma_\theta(xt, t)$, they can fix it to a constant, choosing either $\beta_t\mathbf{I}$ or $\tilde\beta_t\mathbf{I}$. These values correspond to upper and lower bounds for the true reverse step variance.

\section{Observation Guidance}
Here, we supplement the exploitation of observation data as guidance to improve the reconstruction. Specifically, we utilize a conditioner $p_\phi(y|x_t,t)$ and use $\nabla_{x_t}p_\phi(y|x_t,t)$ to guide the diffusion sampling process towards given observation $y$. 
We will describe how to use such conditioners to improve the accuracy of the reconstruction result. The
notation is chosen as $p\phi(y|x_t, t) = p_\phi(y|x_t)$ and
$\epsilon_\theta(x_t, t) = \epsilon_\theta(x_t)$ for brevity, referring to separate functions for each time step $t$.

\subsection{Diffusion Process}
% The complete derivation of guided diffusion is provided by Dhariwal and 181
% Nichol ~\citep{dhariwal2021diffusion}, as well as Ho and Salimans [37]~\citep{ho2022classifier}
According to Bayes' theorem:
\begin{align}
    p(x_t|x_{t+1}, y) & =\frac{p(x_t,x_{t+1},y)}{p(x_{t+1},y)}=\frac{p(x_t|x_{t+1})p(x_{t+1})p(y|x_t,x_{t+1})}{p(y|x_{t+1})p(x_{t+1})}\\ &=\frac{p(x_t|x_{t+1})p(y|x_t,x_{t+1})}{p(y|x_{t+1})}=\frac{p(x_t|x_{t+1})p(y|x_t)}{p(y|x_{t+1})}\\ &\propto p(x_t|x_{t+1})p(y|x_t),
\end{align}
where the penultimate step is derived from :
\begin{equation}
p(y|x_t,x_{t+1})=p(x_{t+1}|x_t,y)\frac{p(y|x_t)}{p(x_{t+1}|x_t)}=p(x_{t+1}|x_t)\frac{p(y|x_t)}{p(x_{t+1}|x_t)}=p(y|x_t).
\end{equation}
The $\hat p(y|x_{t+1})$ can be treated as a constant because $y$ does not depend on $x_{t+1}$. Therefore, we want to sample from the distribution $C\cdot p (x_t|x_{t+1})\hat p(y|x_t)$ where $C$ denotes the normalization constant. $p(x_t| x_{t+1})$ is just the $p_\theta(x_t | x_{t+1})$ approximated by a neural network, so the rest is $\hat p (y | x_t)$ that can be obtained by computing a conditioner (evaluated by the distance between $y$ and $\hat x_0(x_t)$) on noised images
$x_t$.

\subsection{Guided Reverse Process}
Assume a diffusion model with an unconditional reverse noising process is $p_\theta(x_t|x_{t+1})$. In sea temperature reconstruction, we will obtain some sparse observation data and aim to impute the whole temperature field. Therefore, we regarded $y$ as sparse observation data, $x_t$ as the generated reconstruction in time step $t$, $\hat x_0(x_t)$ as the final reconstruction result from $x_t$. Then the conditioner is formulated as follows:
\begin{equation}
    p(y|x_t)=exp(-s\cdot\mathcal{L}(\hat x_0(x_t), y )),
\end{equation}
where $\mathcal{L}$ stands for Euclidean distance between $\hat x_0(x_t)$ and $y$.
We have known that
\begin{equation}
    p_\theta(x_t|x_{t+1},y)=C\cdot p_\theta(x_t|x_{t+1})p(y|x_t),
\end{equation}
where $p_\theta(x_t|x_{t+1})$ is the unconditional reverse process in denoising diffusion probabilistic models, which can be approximated as a perturbed Gaussian distribution $p_\theta(x_t|x_{t+1})=\mathcal{N}(\mu,\Sigma)$:
\begin{equation}
    \log p_\theta(x_t|x_{t+1})=\mathcal{N}(\mu,\Sigma)=-\frac{1}{2}(x_t-\mu)^T\Sigma^{-1}(x_t-\mu) + C.
\end{equation}
We can assume that $\log p(y|x_t)$ owns low curvature
when compared with $\Sigma^{-1}$. This assumption is reasonable under the constraint that the infinite diffusion step, where $\|\Sigma\|\rightarrow 0$. Under the circumstances, $\log p(y|x_t)$ can be
approximated via a Taylor expansion around $x_t = \mu$ as:
\begin{align}
\log p(y|xt) &\approx \log p(y | x_t)|_{x_t=\mu}
+ (x_t-\mu)\nabla x_t\log p(y | x_t)|_{x_t=\mu}\\
&= (x_t-\mu) g + C_1,
\end{align}
where $g =\nabla x_t\log p(y | x_t)|_{x_t=\mu}$, and $C_1$ is a constant. This gives~\citep{dhariwal2021diffusion}:
\begin{align}
    \log(p_\theta(x_t|x_{t+1})p(y | xt))&\approx -\frac{1}{2}(x_t-\mu)^T\Sigma^{-1}(x_t-\mu) + (x_t-\mu) g + C_2 \\ &\approx
    -\frac{1}{2}(x_t-\mu-\Sigma g)^T\Sigma^{-1}(x_t-\mu-\Sigma g) + \frac{1}{2}g^T\Sigma g + C_2 \\ &\approx
    -\frac{1}{2}(x_t-\mu-\Sigma g)^T\Sigma^{-1}(x_t-\mu-\Sigma g) + C_3 \\ &=
    \log p(z) + C_4, z\sim\mathcal{N}(\mu+\Sigma g,\Sigma),
\end{align}
where the constant term $C_4$ could be ignored. In practice, we use $g =\nabla x_0\log p(y|x_t)|_{x_t=\mu}=\nabla x_0\log \mathcal{L}(\hat x_0(x_{t+1}))$ instead, which simplifies the calculation and maintains great performance.

\begin{figure}[t]
    \centering
    \includegraphics[width=1\linewidth]{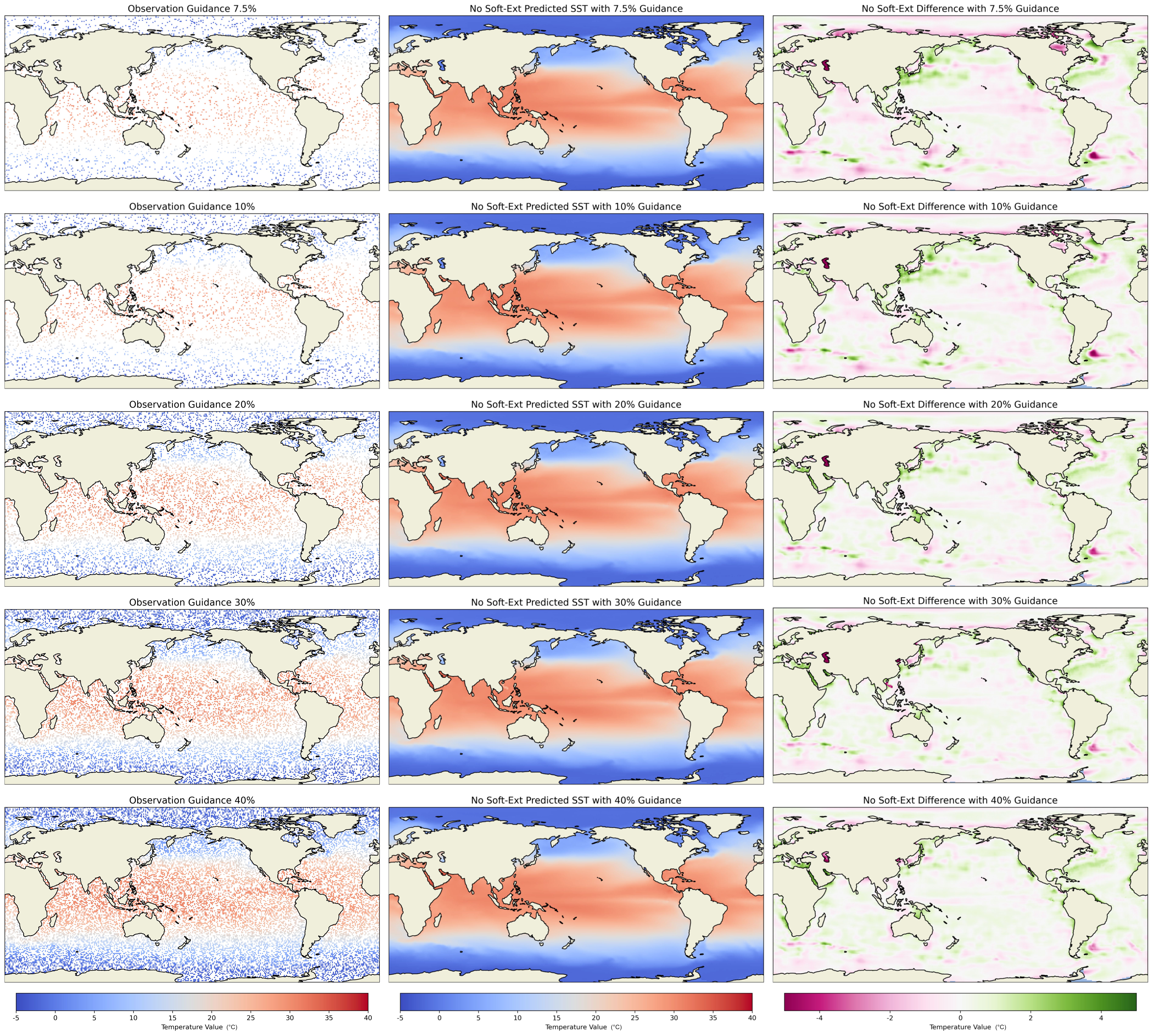}
    \caption{Comparison of increasing observation density with no soft extension, separately 7.5\%, 10\%, 20\%, 30\%, 40\% from top to bottom.}
    \label{fig:guide-no-soft}
\end{figure}

\section{Visualization of Scalability Under Increasing Observation Density}
\begin{figure}[tbp]
    \centering
    \includegraphics[width=1\linewidth]{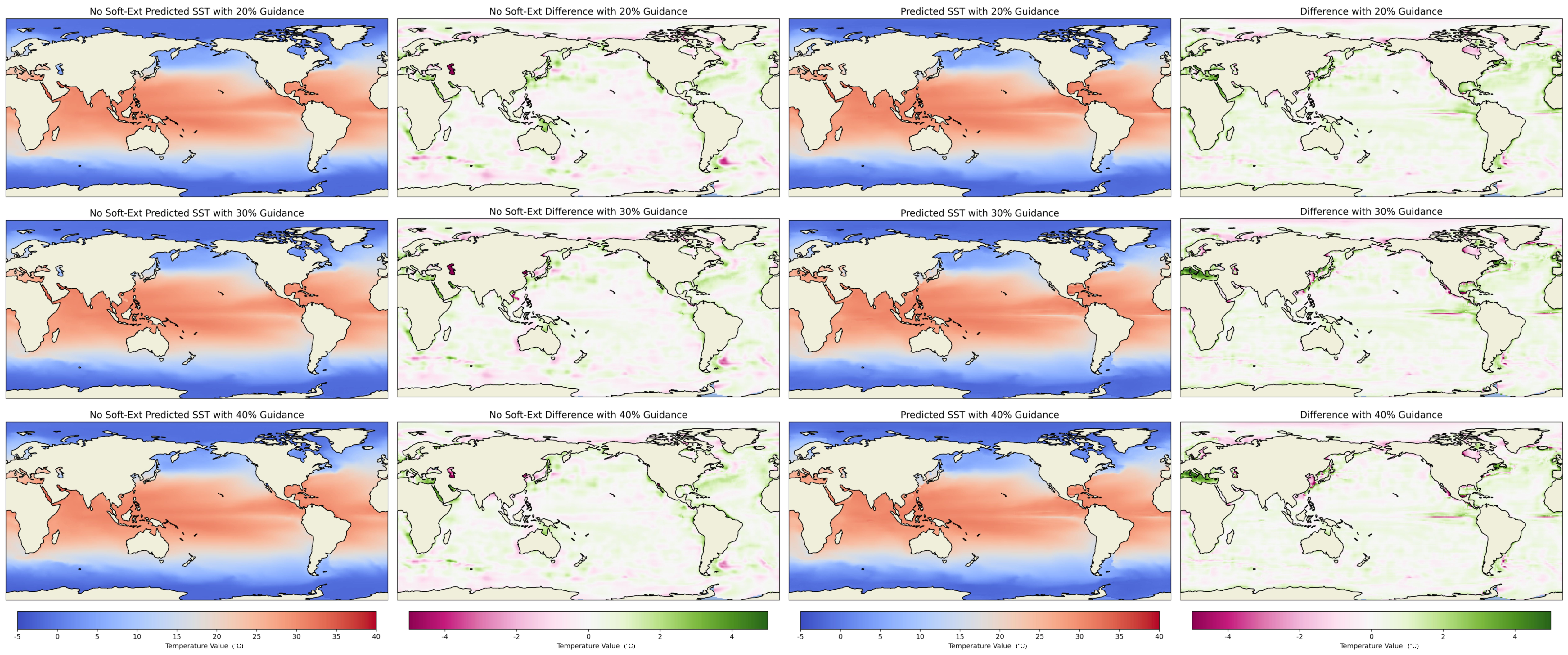}
    \caption{Comparison of complete method(right) and method with no soft extension at dense observation density(left), separately 20\%, 30\%, 40\% from top to bottom.}
    \label{fig:compare-guide}
\end{figure}

We also visualize the reconstruction result of different guided rates to show the scalability under increasing observation density.
We evaluate two variants: ReconMOST with and without the soft-extension module. 
% We conduct experiments using our ReconMOST and a variant ReconMOST without soft-extension separately. 
Fig.~\ref{fig:guide-no-soft} demonstrates the reconstruction result without soft-extension, demonstrating that higher guidance density leads to improved accuracy in reconstruction.
% as the guidance density increases, the reconstruction result achieved more accurate.
However, the soft-extension may cause variance in the equatorial region and loss of otherwise good physical consistency when guidance density increases, as shown in Fig.~\ref{fig:compare-guide}. 
% 观测点足够多
This occurs because sufficiently dense observations enable accurate temperature field reconstruction through guided reverse sampling alone, without requiring soft-extension.
Adding extra information via a Gaussian kernel will destroy the learned local physical distribution underlying the pre-trained model, but most areas performed better than that without soft-extension.
% 7.5 10 20 30 40 50
% （原图）指导比例，重建效果，差距 3*3为适宜，4*4太小

\section{Visualization Across Temporal Dimension}
\begin{figure}[htp]
    \centering
    \includegraphics[width=1\linewidth]{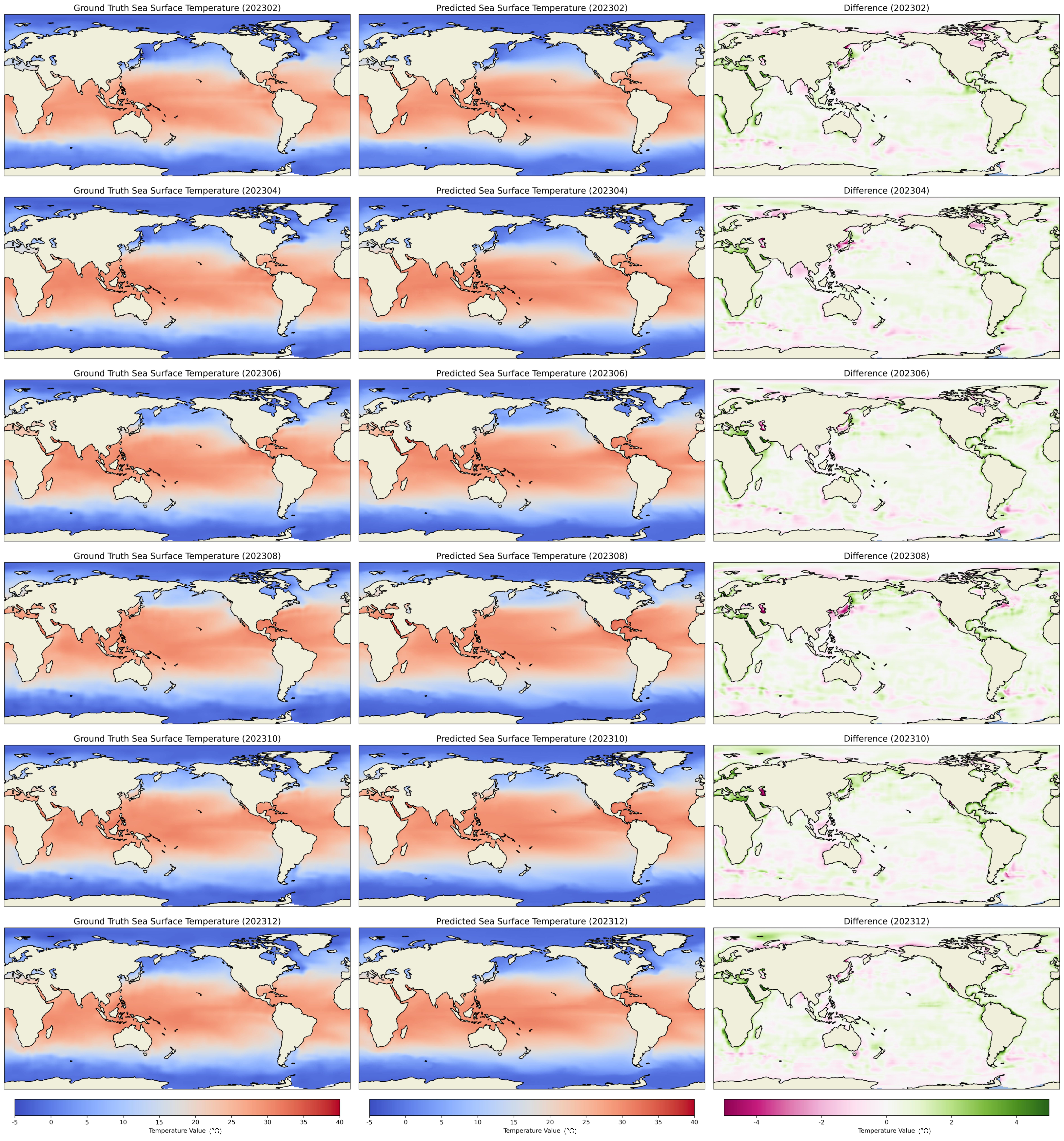}
    \caption{Enter Caption}
    \label{fig:diff-time}
\end{figure}

We chose different timestamps (every two months in 2023) to indicate the temporal compatibility of our ReconMOST. 
As shown in Fig.~\ref{fig:diff-time}, our method achieved excellent reconstruction performance at different times, only April and August have relatively obvious local errors.
These results confirm the temporal stability of our method and support its potential for temporal extension in future work.
% , which is a good verification for extending our method to time dimension in the future.

% 时间尺度的比较，4-6个时刻，最好的模型，比如2023年2 4 6 8 10
% 原图，重建效果，差距

% \section{Accuracy of Reconstruction in Single Layer}
% In the paper, we shows average MSE of top ten layers. Right there we shows the specific accuracy of reconstruction for each experiment. 
% % 待定，有点长，考虑和下一部分合并

\section{Compared with Vanilla U-Net Method}
To evaluate the performance, robustness and adaptability of our ReconMOST for ocean temperature reconstruction, we compared it with a vanilla U-Net model. 
The architecture of U-Net here is identical to that we used for guided diffusion, ensuring the same number of model parameters. 
We pre-trained U-Net models with Top5 CMIP6 simulations ranked by MSE performance as defined in our Experiments and Analysis section. 
We evaluated model performance on the EN4 dataset from the Met Office.
Results here in Table~\ref{tab:table5} show that our ReconMOST outperforms U-Net, with lower reconstruction loss and requiring fewer training steps. 
The performance gap here can be attributed to a distribution mismatch that exists between the CMIP6 model simulations and real-world EN4 analysis data, a common challenge in applying machine learning to physical systems~\citep{wilson2020,furner2022sensitivity}. 
Unlike U-Net, our ReconMOST takes partial real-world observations to adjust its distributions, improving its accuracy and adaptability for sea temperature reconstruction. 
The percentage of observation points during pre-training that affects reconstruction quality is another point worth considering, and will be left for future work to discuss.

\begin{table}[t]
    \centering
    \caption{UNet reconstruction result of EN4 analysis with difference guided rate (number of guided points): use best model Top5 in Table~\ref{tab:table5}. Guided rate during pre-training and number of training steps are displayed. The last three columns are, respectively, average MSE of upper ten levels on guidance points (MSE-g), reconstruction points (MSE-r), and all complete global points (total MSE).}
    \vspace{+0.2cm}
    % \caption{Guided Reconstruction Result of EN4 Analysis with Difference Guided Rate(number of guided points): Use best model Top5 in table~\ref{tab:table2} and other parameters s=4, Sigma=0, soft-ext size is 5.The last three columns are respectively average MSE of upper ten levels on guidance points(MSE-g), reconstruction points(MSE-r) and all complete global points(total MSE).}
    % 定义表格列格式：
    % | 表示竖线，l 表示左对齐，c 表示居中，r 表示右对齐
    % 竖线放在 mode 和 s、softmask 和 loss1 之间
    % \resizebox{\textwidth}{!}{
    \begin{tabular}{c c c c c c}
        \toprule
        % 表格标题行
        \small\textbf{pre-train guided rate} & \small\textbf{guided rate} &
        \small\textbf{\# training steps} & 
        \small\textbf{MSE-g} & \small\textbf{MSE-r} & \small\textbf{total MSE} \\
        \midrule
        % 数据行（示例数据）
        7.5\%&10\% & 200,000 & 7.975 & 0.870 & 7.264 \\
        7.5\%&10\% & 800,000 & 4.946 & 0.530 & 4.504 \\
        7.5\%&40\% & 200,000 & 11.247 & 7.414 & 9.714 \\
        7.5\%&40\% & 800,000 & 3.683 & 2.408 & 3.173 \\
        \midrule
        40\%&10\% & 200,000 & 9.102 & 1.223 & 8.314 \\
        40\%&10\% & 800,000 & 9.816 & 1.362 & 8.970 \\
        40\%&40\% & 200,000 & 6.022 & 4.011 & 5.218 \\
        40\%&40\% & 800,000 & 6.305 & 4.198 & 5.462 \\
        \bottomrule
    \end{tabular}
    % }
    \label{tab:table5}
\vspace{-0.4cm}
\end{table}
% 缺一个wo，不加任何改进的

% \section{Limitation and Future Work}
% 对于没有指导点的内海区域重建效果不好
% 观测点之间的关系利用不充分（数据同化）
% \input{table6}

\end{document}